\documentclass[conference]{IEEEtran}
\IEEEoverridecommandlockouts
\usepackage{cite}
\usepackage{amsmath,amssymb,amsfonts}
\usepackage{algorithmic}
\usepackage{graphicx}
\usepackage{textcomp}
\usepackage[table]{xcolor}
\usepackage{parskip}  
\usepackage[caption=false,font=footnotesize]{subfig}
\usepackage{multirow}
\usepackage{algorithm}
\usepackage[top=72pt,bottom=54pt,right=54pt,left=54pt]{geometry}

\newcommand{\norm}[1]{\left\lVert#1\right\rVert}
\newcommand{\argmin}[1]{\underset{#1}{\mathrm{argmin}}}

\DeclareMathOperator*{\Circ}{\circ}

\begin{document}

\title{Multi-View Graph Embedding Using Randomized Shortest Paths \\ 
\thanks{This work was supported by the US National Science Foundation (NSF) grants CCF-1319653 and CCF-1553075.}
}

\author{
\IEEEauthorblockN{Anuththari Gamage\IEEEauthorrefmark{1}, Brian Rappaport\IEEEauthorrefmark{1}, Shuchin Aeron\IEEEauthorrefmark{1}, and Xiaozhe Hu\IEEEauthorrefmark{2}}
\IEEEauthorblockA{\IEEEauthorrefmark{1}Department of Electrical and Computer Engineering,
Tufts University\\
anuththari.gamarallage@tufts.edu \qquad brian.rappaport@tufts.edu \qquad shuchin@ece.tufts.edu }
\IEEEauthorblockA{\IEEEauthorrefmark{2}Department of Mathematics,
Tufts University\\
xiaozhe.hu@tufts.edu }
}

\maketitle

\begin{abstract}
Real-world data sets often provide multiple types of information about the same set of entities. This data is well represented by multi-view graphs, which consist of several distinct sets of edges over the same nodes. These can be used to analyze how entities interact from different viewpoints. Combining multiple views improves the quality of inferences drawn from the underlying data, which has increased interest in developing efficient multi-view graph embedding methods. We propose an algorithm, C-RSP, that generates a common (C) embedding of a multi-view graph using Randomized Shortest Paths (RSP). This algorithm generates a dissimilarity measure between nodes by minimizing the expected cost of a random walk between any two nodes across all views of a multi-view graph, in doing so encoding both the local and global structure of the graph. We test C-RSP on both real and synthetic data and show that it outperforms benchmark algorithms at embedding and clustering tasks while remaining computationally efficient.
\end{abstract}


\section{Introduction}

To model and understand complex systems, we must consider how different entities within a system relate to one another. Many such relational data sets provide multiple views of the same underlying set of entities. For example, a group of people can be characterized by their interactions with one another on a social networking platform. As a first approximation, we can consider only whether a relationship exists between two people on this platform. However, we can also study their interactions across other platforms, or across the different modes of communications provided by the platform \cite{MultiviewTwitter}, which provide us multiple views of the relationships between the same group of people. Other examples of multi-view data sets include multi-omics measurements in single cell RNA sequencing data \cite{COLOMETATCHE2018} and 2D projections of a single 3D object captured from multiple angles for 3D reconstruction \cite{Kolev:2009}. 

Graphs are used extensively to model this type of relational data for machine learning tasks, where the nodes or vertices of the graph represent the entities studied in the data set and edges represent their relationships. By learning a vector for each node in the graph, we can create a graph embedding, which gives the entities in the data set a representation in Euclidean space. These embeddings can be used for various applications such as data visualization, clustering, and link prediction. There are several well-known algorithms for creating graph embeddings from a single-view graph. The extension to multi-view or multi-layer graphs, however, has not been well studied up to this point. Providing more views leads to improved accuracy in clustering and embedding. This has increased recent interest in developing efficient multi-view graph embedding methods.

In a multi-view graph embedding, each node is assigned a vector that incorporates data from all views of the graph. Simple methods to create multi-view embeddings include combining multiple views of the graph into one graph using an AND/OR aggregation of the edge sets and embedding the resulting single graph, or embedding each view independently and concatenating the different embeddings obtained for each node \cite{mvn}. More sophisticated algorithms have been developed based on matrix factorization \cite{LMF, MultiNMF}, tensor factorization \cite{Tucker2, graphFUSE}, and spectral embedding \cite{SCML, Coreg, Cotraining}. Many of these algorithms focus on clustering multi-view graphs, a specific application thereof. High clustering accuracy indicates a good embedding since relative similarity between nodes should be correctly reflected in the embedding.

The similarity between nodes of a graph can be quantified by a distance measure, such as the shortest path or geodesic distance (number of edges in the shortest path connecting two nodes) or the commute time distance (expected number of edges in a random walk from one node to the other and back). If two nodes are similar, then they are likely to have a shorter distance between them. Ideally, graph embeddings should preserve the distances between the nodes in their respective node embeddings. The commute time distance encodes a graph's clusters better than the shortest path distance \cite{VonLuxburg:2014, Luxburg:Spectral_Clustering}. However, for large graphs with greater than 1000 nodes, or for graphs where the dimensionality of the underlying data is high, the commute time distance fails to capture the global structure of the graph accurately \cite{VonLuxburg:2014:HCT}. This is because it degenerates to a function of the node degrees, which captures only the local connectivity of the nodes \cite{VonLuxburg:2014:HCT}. 

\newgeometry{top=54pt,bottom=56pt,right=54pt,left=54pt}

In light of these deficiencies of the commute time and shortest path distance measures, there has been increased interest in alternative distance measures that generalize these two distances \cite{Hashimoto:2015, Yen:2008, Kivimaki}. The Randomized Shortest Path (RSP) dissimilarity measure \cite{Yen:2008} generalizes the two distances by computing an intermediate measure parameterized by a tunable variable $\beta$, such that both limiting cases reduce to one of these measures: as $\beta \to \infty$, the RSP dissimilarity reduces to the shortest path distance and as $\beta \to 0$, it reduces to the commute time distance. This type of distance measure is particularly suitable for graph embedding as it preserves in the embedding space both the local and global features of the manifold from which the data set is sampled. 

In this paper, we propose a generalized distance on multi-view graphs called the \textbf{Common Randomized Shortest Path Dissimilarity (C-RSP)} based on the RSP dissimilarity on single-view graphs. We highlight some of the advantages of the proposed approach below:
\begin{enumerate}
\item Like RSP, C-RSP generalizes the shortest path and commute time distances on multi-view graphs using a single parameter $\beta$. As $\beta\to\infty$, it reduces to the shortest path distance and as $\beta\to0$, it reduces to the commute time distance. This type of generalized distance generates more accurate graph embeddings, and as a result, produces higher clustering and visualization accuracy for a given data set.

\item The RSP dissimilarity has an intuitive interpretation as the minimum expected cost of a random walk between any two nodes of a graph over all possible transition probability matrices \cite{Yen:2008}. C-RSP has a similar interpretation: the minimum expected cost of a random walk between nodes across all views.

\item C-RSP retains the computational efficiency of RSP \cite{Yen:2008, Kivimaki}. The proposed algorithm first combines the multiple views and then generates an embedding using the combined matrix, eliminating the need for factorization and simultaneous optimization approaches found in other multi-view graph embedding algorithms. This makes it less computationally intensive and more scalable.
\end{enumerate}

The rest of the paper is organized as follows: Section II provides an overview of the Randomized Shortest Path dissimilarity measure. Section III describes the proposed C-RSP algorithms and its derivation. Section IV presents experimental results comparing C-RSP to benchmark multi-view algorithms on a variety of synthetic and real-world data sets, comparing their clustering and embedding accuracy. Section V discusses our results and future work. 

\section{Randomized Shortest Path Dissimilarity}

\subsection{Mathematical Preliminaries}
Let $G = \{V,E\}$ be a simply connected graph, where $V = \{1, \ldots, n\}$ denotes the set of nodes of the graph and $E = \{(i,j)\,|\, i,j \in V\}$ denotes the set of edges between nodes. This graph can be represented by its affinity matrix $A \in \mathbb{R}^{n \times n}$, where the elements $a_{ij}$ are termed the \emph{affinities} or \emph{weights}. $a_{ij} = 1$ for $(i,j)\in E$ in unweighted graphs, $a_{ij} \ne 0$ for $(i,j)\in E$ in weighted graphs, and for all graphs $a_{ij} = 0$ if $(i,j)\notin E$. The degree matrix $D\in \mathbb{R}^{n \times n}$ is a diagonal matrix containing the weighted degree (the sum of all edges leaving the node) of node $i$ in element $D_{ii}$ and zeros elsewhere.

We can compute the transition probability matrix $P^{\mathrm{ref}} = D^{-1}A$ of the graph $G$, which is row-stochastic and defines a probability distribution on the edges of the graph. A random walk on the graph follows a sequence of nodes with the order determined by these transition probabilities. Consider a particular path on this graph starting at a source node $s$ and a destination node $t$, denoted by $p_{s\rightarrow t} = \{s, v_1, v_2, \ldots , v_m, t\}$. Then the probability of the path is given by the product $P^{\mathrm{ref}}_{s,v_1}P^{\mathrm{ref}}_{v_1,v_2}\dots P^{\mathrm{ref}}_{v_m,t}$, denoted $P^{\mathrm{ref}}(p_{s\rightarrow t})$.

Since affinities refer to a positive correlation between nodes, we define the \emph{cost} of each edge $(i,j)$ by a cost matrix $C$ with elements $c_{ij} = a_{ij}^{-1}$ where $0 < c_{ij} < \infty$. We can compute the total cost for a given path $p_{s\rightarrow t}$, denoted by $C(p_{s\rightarrow t}) = c_{s,v_1} + c_{v_1, v_2} + \ldots + c_{v_m,t}$.

An \emph{absorbing} path is a path where the destination node $t$ has no outgoing edges except to itself ($c_{t,t} = 1, c_{t,k} = \infty\,, k\neq t \in V$). The cost of an absorbing path (even permitting infinite length ones) is finite since a random walk on the path will terminate in a finite number of steps with probability 1. 

For C-RSP, we consider only absorbing paths from $s$ to $t$, and our path is denoted $p_{s\to t}$, with the path probability under a given distribution $P$ denoted by $P(p_{s\to t})$ and the cost of traversing the path denoted by $C(p_{s\to t})$. Suppose the set of all such absorbing paths is $\mathcal{P}_{s\to t}$. Then, the \emph{expected cost} of a random walk from a source node $s$ to a destination node $t$ over a given distribution $P$ is given by $\sum\limits_{p \in\mathcal{P}_{s\to t}} P(p)C(p)$.

\subsection{Randomized Shortest Paths Dissimilarity}\label{section:rsp_deriv}

The Randomized Shortest Path (RSP) is defined to be the path between two nodes with the minimum expected cost over all transition probability matrices \cite{Yen:2008}. In order to constrain a random walk between two nodes to a RSP, we compute a new probability distribution $P(p)$ that achieves the minimum expected cost among all possible probability distributions having  fixed relative entropy (K\"ullback-Leibler divergence) with respect to the reference probability distribution $P^{\mathrm{ref}}(p)$: 

\vspace{-3pt}
\begin{equation}
\begin{split}
& P^{RSP} = \argmin{P}{\sum\limits_{p \in \mathcal{P}_{s\to t}} P(p)C(p)} \\
& \text{subject to }\sum\limits_{p \in \mathcal{P}_{s\to t}}P(p)\ln\frac{P(p)}{P^{\mathrm{ref}}(p)} = J_0, \\
& \: \qquad \qquad \sum\limits_{p \in \mathcal{P}_{s\to t}} P(p) = 1
\end{split}
\end{equation}

\newpage

The solution to this constrained optimization is given by the following expression for any $p_{s\to t} \in \mathcal{P}_{s\to t}$ \cite{Yen:2008}:
\begin{equation}
\label{rsp_eqn}
P^{RSP}(p_{s\to t}) = \frac{P^{\mathrm{ref}}(p_{s\to t})e^{-\beta C(p_{s\to t})}}{\sum\limits_{p\in \mathcal{P}_{s\to t}}P^{\mathrm{ref}}(p)e^{-\beta C(p)}} 
\end{equation}

Using the probability distribution for Randomized Shorted Paths derived above, we can define the symmetric RSP dissimilarity between two nodes as follows. Suppose the expected cost of traversing the randomized shortest path between source node $s$ and destination node $t$ is given by 

$$\overline{C}_{st} = \sum\limits_{p\in\mathcal{P}_{s\to t}} P^{RSP}(p)C(p).$$

This expression is not guaranteed to be symmetric, so we calculate the symmetric RSP dissimilarity measure between the two nodes by

\begin{equation}
\Delta^{RSP}_{st} = \frac{\overline{C}_{st} + \overline{C}_{ts}}{2}.
\end{equation}

Note that the computed RSP distance measure is termed a ``measure" instead of a ``metric" since it does not follow the triangle inequality for certain ranges of $\beta$ used \cite{Yen:2008, Kivimaki}.

\subsection{Efficient Computation of the RSP Dissimilarity}

Following the derivation of the RSP dissimilarity by Yen et al.\cite{Yen:2008}, an efficient closed-form expression for its computation was derived by Kivim\"aki et al.\cite{Kivimaki}, which we describe in Algorithm \ref{alg:rsp}. This computation is done entirely through matrix operations, which lends itself nicely to input graphs in the form of affinity matrices. The output of the algorithm is the symmetric matrix $\Delta^{RSP} \in \mathbb{R}^{n\times n}$, in which each entry $\Delta^{RSP}_{ij}$ gives the RSP dissimilarity between the nodes $i$ and $j$.  
\begin{algorithm}
    \caption{RSP Dissimilarity}
    \label{alg:rsp}
    \begin{algorithmic}
    	\REQUIRE $A \in \mathbb{R}^{n\times n}$ (affinity matrix for a simply connected graph $G$), $\beta$ (optimization parameter)
    	\ENSURE $\Delta^{RSP} \in \mathbb{R}^{n\times n}$ (RSP dissimilarity matrix)
    	\vspace{5pt}
    	\STATE  $P^{\mathrm{ref}} = D^{-1}A$  ($D$ is the degree matrix of $A$)
    	\STATE $C = 1 \div A$ (element-wise inverse)
    	\STATE $ W = P^{\mathrm{ref}} \circ e^{ -\beta C}$ (element-wise multiplication)
    	\IF{$\rho(W) \geq 1$}
    	    \STATE Stop: will not converge
        \ENDIF	
    	\STATE $Z = (I - W)^{-1}$   $\qquad (I \in \mathbb{R}^{n \times n}$ is the identity matrix)
    	\STATE $S = (Z[C\circ W]Z)\div Z$
    	\STATE $\overline{C} = S - \mathbf{1}d_S^T$   $\qquad  (\mathbf{1}, d_S \in\mathbb{R}^n$, $d_i = S_{ii}$)
    	\STATE $\Delta^{RSP} = \frac{1}{2}(\overline{C} + \overline{C}^T)$
    \end{algorithmic}
\end{algorithm}

\section{Combining Multiple Graph Views Using Common Randomized Shortest Paths}

\subsection{Deriving a Common RSP Probability Distribution}
In this work, we extend the core RSP framework to generate a multi-view graph distance measure. If we represent a single-view graph by $G = \{V,E\}$, then a multi-view graph is denoted $\mathcal{G} = \{V, (E_1, \ldots, E_m)\}$ where each view is given by $G_i = \{V, E_i\}$. We represent this graph with an $n\times n\times m$ affinity tensor, where each $n\times n$ slice of the tensor $A_i$ represents the affinity matrix for that edge set. Note that each $G_i$ is assumed to be a simply connected graph.

We first derive a common probability distribution, $P^{CRSP}$, over all views of the graph. This is accomplished by minimizing the expected cost for all possible paths on all views, with the condition that the common distribution $P^{CRSP}$ and the reference probability distribution of each view, $P_i^{\mathrm{ref}}$, have the same fixed relative entropy. This constrained optimization is represented as follows, with reference probability distributions $P_1^{\mathrm{ref}},\dots,P_m^{\mathrm{ref}}$ and cost matrices $C_1,\dots,C_m$: 
\begin{equation}
\begin{split}
& P^{CRSP} = \argmin{P}{\sum_{i=1}^{m}\sum_{p\in\mathcal{P}_{s\to t}}P(p)C_i(p)} \\
& \text{subject to } \sum_{i=1}^m\sum_{p\in\mathcal{P}_{s\to t}}P(p)\ln\frac{P(p)}{P_i^{\mathrm{ref}}(p)} = J_0, \\
& \!\! \qquad \qquad \sum\limits_{p \in \mathcal{P}_{s\to t}}P(p) = 1
\end{split}
\end{equation}

Solving this constrained optimization results in the following probability distribution, which we term the Common Randomized Shortest Paths (C-RSP) distribution. Multiplication indicated by ``$\circ$'' is done element-wise. The full derivation of this expression is detailed in the Appendix. This equation holds for any $p_{s\to t} \in \mathcal{P}_{s\to t}$.

\begin{equation}
\label{crsp_eqn}
P^{CRSP}(p_{s\to t}) = \frac{\sqrt[m]{\Circ\limits_{i=1}^{m}P_i^{\mathrm{ref}}(p_{s\to t})}\circ e^{-\beta \sum\limits_{i=1}^{m}C_i(p_{s\to t})}}{\displaystyle\sum_{p\in\mathcal{P}_{s\to t}}\sqrt[m]{\Circ\limits_{i=1}^{m}P_i^{\mathrm{ref}}(p)}\circ e^{-\beta\sum\limits_{i=1}^{m} C_i(p)}}
\end{equation}

\subsection{Common Randomized Shortest Paths Dissimilarity}

Using the derived common probability distribution, $P^{CRSP}$, we can compute a dissimilarity measure $\Delta^{CRSP}$ for multi-view graphs following an approach similar to that detailed in section \ref{section:rsp_deriv} above. Note that we can use the same algorithm used for computing RSP if we were to have a single reference probability matrix and a single cost matrix instead of the tensors associated with a multi-view graph. Using the expression for $P^{CRSP}(p_{s\to t})$ derived in equation \eqref{crsp_eqn}, we can combine the individual views of these tensors to obtain these matrices as detailed below.

Let $\overline{\mathbf{P}}$ denote the combined reference transition probability matrix and $\mathbf{\overline{C}}$ denote the combined cost matrix. By comparing equations \eqref{rsp_eqn} and \eqref{crsp_eqn}, we obtain

\begin{equation}
\label{combine_p}
\mathbf{\overline{P}} = \frac{\sqrt[m]{\Circ\limits_{i=1}^{m}P_i^{\mathrm{ref}}}}{\mathbf{1}^T\cdot\sqrt[m]{\Circ\limits_{i=1}^{m}P_i^{\mathrm{ref}}}}
\end{equation}

\begin{equation}
\label{combine_c}
\mathbf{\overline{C}} = \sum\limits_{i=1}^{m} C_i
\end{equation}

Note that in equation \eqref{crsp_eqn}, we derive the probability of an individual path and not the entire set of possible paths. Thus, we need to take care to omit instances when the path does not exist given a particular $P_i^{\mathrm{ref}}$, which occurs when an entry in any $P_i^{\mathrm{ref}}$ is zero. Also note that the multiplication in this expression is taken element-wise, as is the $m^{th}$ root. Finally, note that this manner of combining the different $P_i^{\mathrm{ref}}$ matrices does not guarantee a row-stochastic matrix, which is necessary for it to be a probability distribution. Thus, the resulting matrix $\mathbf{\overline{P}}$ must be normalized to obtain a row-stochastic matrix. This can be achieved easily by dividing the entries in each row by the row sum. In this way, we obtain a combined reference probability matrix $\mathbf{\overline{P}}$ and a combined cost matrix $\mathbf{\overline{C}}$ that can then be used in the original RSP algorithm to obtain the C-RSP dissimilarity measure $\Delta^{CRSP}$, as detailed in Algorithm~\ref{alg:crsp}. 

\begin{algorithm}
    \caption{C-RSP Dissimilarity}
    \label{alg:crsp}
    \begin{algorithmic}
    	\REQUIRE $A_1, \ldots, A_m$ ($A_i\in \mathbb{R}^{n\times n}$ is the affinity matrix for view $G_i$ of a multi-view graph $G$ where $G_i$ is connected), $\beta$ (optimization parameter)
    	\ENSURE $\Delta^{CRSP} \in \mathbb{R}^{n \times n}$ (C-RSP dissimilarity matrix)
    	\FOR {$i = 1 \ldots m$}
    	\STATE  $P_i^{\mathrm{ref}} = D_i^{-1}A_i$  ($D_i$ is the degree matrix of $A_i$)
    	\STATE $C_i = 1 \div A_i$ (element-wise division)
    	\ENDFOR
    	\STATE $\mathbf{\overline{P}} =  \{ P_1^{\mathrm{ref}}, \ldots, P_m^{\mathrm{ref}}  \}$ combined as given in equation \eqref{combine_p}
    	\STATE $\mathbf{\overline{C}} =  \sum\limits_{i=1}^{m} C_i$ 
    	\STATE $ W = \mathbf{\overline{P}} \circ e^{ -\beta \mathbf{\overline{C}}}$ (element-wise multiplication)
    	\IF{$\rho(W) \geq 1$}
    	    \STATE Stop: will not converge
        \ENDIF	
    	\STATE $Z = (I - W)^{-1}$   $\qquad (I \in \mathbb{R}^{n \times n}$ is the identity matrix)
    	\STATE $S = (Z[C\circ W]Z)\div Z$
    	\STATE $\overline{C} = S - \mathbf{1}d_S^T$   $\qquad  (\mathbf{1}, d_S \in\mathbb{R}^n$, $d_i = S_{ii}$)
    	\STATE $\Delta^{CRSP} = \frac{1}{2}(\overline{C} + \overline{C}^T)$
    \end{algorithmic}
\end{algorithm}

Using this C-RSP dissimilarity matrix, we obtain a multi-view graph embedding by applying Multidimensional Scaling. To cluster a multi-view graph, we can use Spectral Clustering~\cite{Luxburg:Spectral_Clustering} on $(\Delta^{CRSP})^{-1}$, which serves as a measure of affinity.

\section{Experimental Results}

For a general graph, evaluating the effectiveness of its embedding is not straightforward, as we do not know \emph{a priori} how distant the nodes should be once they are embedded or how they should be oriented relative to each other. However, if we know that the graph contains latent clusters of nodes, we can assume that the nodes belonging to the same cluster are likely to appear closer together in their embedding, while those that are in different groups are likely to be distant. Thus, for data sets with latent clusters, we can evaluate the embedding of its representative graph by the clustering accuracy achieved using the embedding vectors.

In order to evaluate C-RSP, we first test the quality of its embedding on a standard Swiss roll data set and visually compare it to the embeddings generated by other algorithms. We then test its clustering performance against a number of benchmark multi-view graph clustering algorithms on a variety of data sets with latent clusters. Clustering performance is compared using two metrics: the Correct Classification Rate (CCR) as a percent and the Normalized Mutual Information (NMI) between the ground truth and the derived clusters. Our experimental results are available online at https://github.com/Anu-Gamage/C-RSP.

\subsection{Benchmark Algorithms}

\begin{itemize}
\item \textbf{SC-ML}: Spectral Clustering on Multi-Layer Graphs using the Grassmannian Manifold~\cite{SCML} \\
This algorithm embeds a multi-view graph by projecting each of the different views of the graph into the Grassmannian manifold. These projections are then combined into a consensus matrix to represent the full multi-view graph, which is clustered using Spectral Clustering~\cite{Luxburg:Spectral_Clustering}. We use $\lambda = 0.5$ in our tests.

\item \textbf{CSC}: Co-regularized Spectral Clustering~\cite{Coreg} \\
This method combines the views of a multi-view graph using co-regularization, a process that chooses the optimal embedding based upon its similarity with all different views of the data. Two co-regularization algorithms are commonly used: pairwise and centroid-based co-regularization. In this work, we use centroid-based co-regularization, which pushes the eigenvector matrices of all views towards a common consensus matrix. The resulting matrix is then clustered using Spectral Clustering. We use $\lambda = 0.05$ across all tests.

\item \textbf{MultiNMF}: Joint Non-negative Matrix Factorization \cite{MultiNMF} \\
This algorithm factorizes each view of the graph into a basis matrix and a coefficient matrix and computes a consensus matrix such that the coefficient matrices are relatively similar to the consensus matrix. The $i^{th}$ row of the consensus matrix is then taken as the vector embedding of the $i^{th}$ node and clustered using $k$-means clustering. We use the parameters listed in the original code in all of the tests.
\end{itemize}

\begin{figure*}[t]
    \centering
    \begin{minipage}{0.45\textwidth}
        \includegraphics[width=\textwidth]{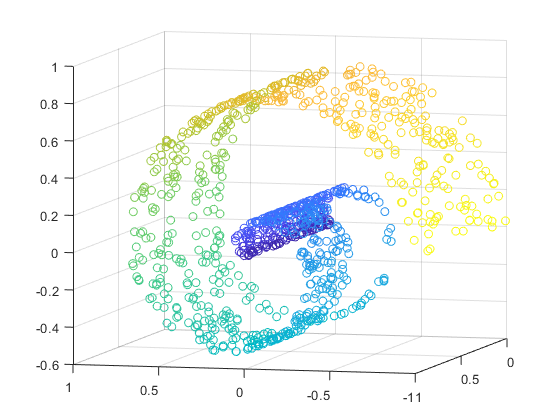}
        \label{fig:sr_gt}
        \caption{Swiss roll ground truth (in 3D), with two holes in the blue and green areas.}
    \end{minipage}
    \hfill
    \begin{minipage}{0.45\textwidth}
        \subfloat[View 1]{
            \includegraphics[width=0.45\textwidth]{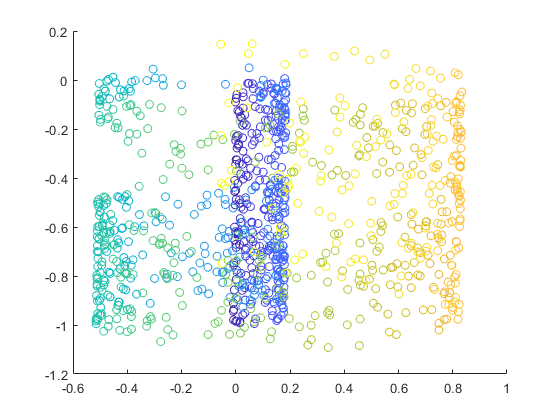}
            \label{fig:sr_v1}
        }
        \subfloat[View 2]{
            \includegraphics[width=0.45\textwidth]{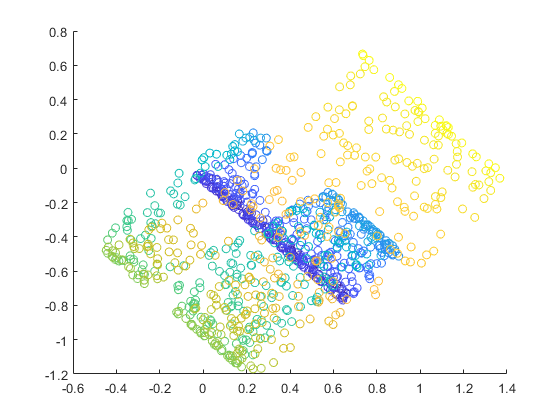}
            \label{fig:sr_v2}
        }
        \linebreak
        \subfloat[View 3]{
            \includegraphics[width=0.45\textwidth]{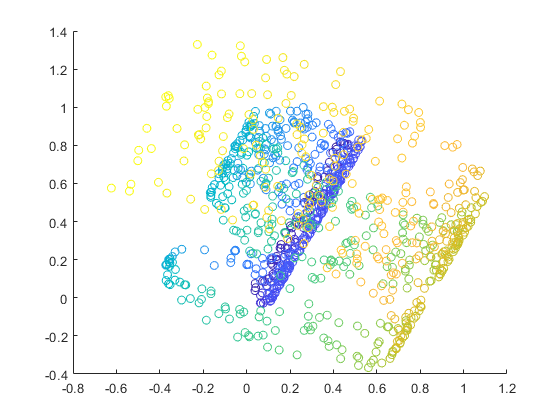}
            \label{fig:sr_v3}
        }
        \subfloat[View 4]{
            \includegraphics[width=0.45\textwidth]{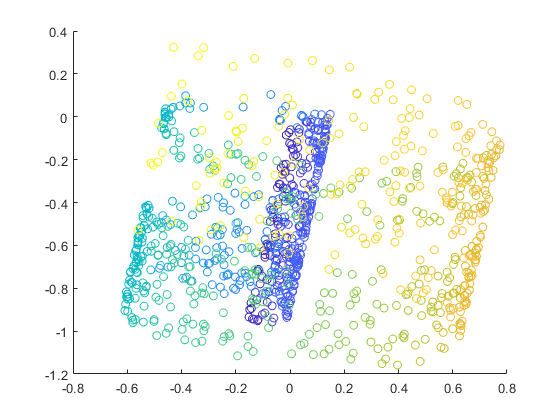}
            \label{fig:sr_v4}
        }
        \caption{Multiple 2D projections of the Swiss roll, used as different views in the multi-view graph.}
    \end{minipage}
    \begin{minipage}{\linewidth}
        \centering
        \subfloat[C-RSP embedding]{
            \includegraphics[width=0.23\textwidth]{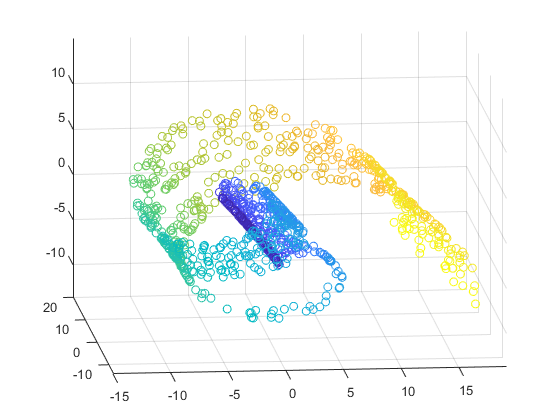}
            \label{fig:sr_crsp}
        }
        \subfloat[SC-ML embedding]{
            \includegraphics[width=0.23\textwidth]{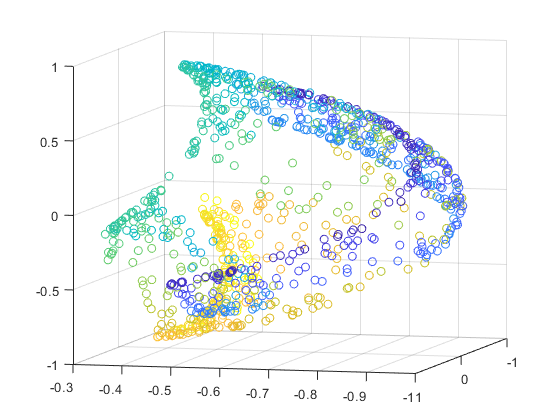}
            \label{fig:sr_scml}
        }
        \subfloat[CSC embedding]{
            \includegraphics[width=0.23\textwidth]{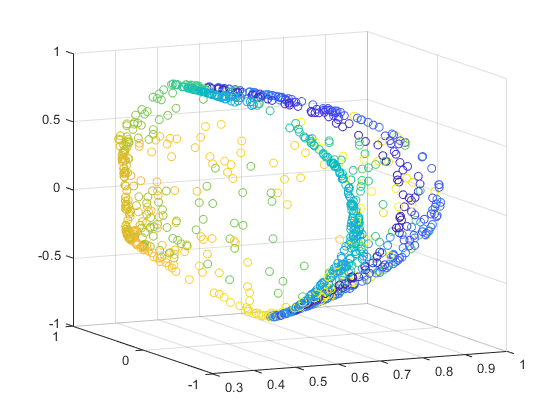}
            \label{fig:sr_csc}
        }
        \subfloat[MultiNMF embedding]{
            \includegraphics[width=0.23\textwidth]{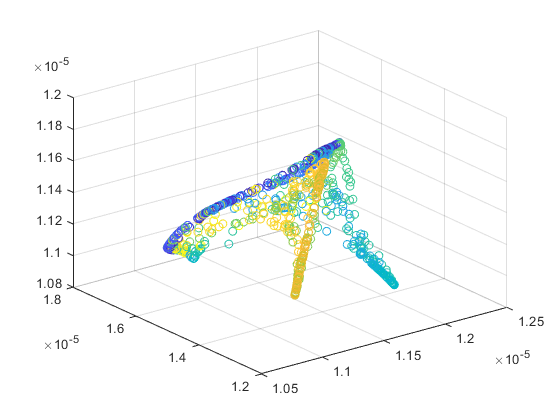}
            \label{fig:sr_mnmf}
        }
        \caption{Embeddings of the Swiss roll generated by C-RSP and other benchmark algorithms. C-RSP retains the Swiss roll shape as well as the relative distances between nodes. SC-ML and CSC both use spectral embedding methods which preserve the relative distances of the nodes but lose the overall structure, while MultiNMF uses a nonnegative matrix factorization to obtain the embedding vectors.}
        \label{fig:sr_embedding}
    \end{minipage}
\end{figure*}

\subsection{Synthetic Data Sets}
\begin{itemize}
    \item \textbf{Swiss Roll}:
    To compare the embedding performance across algorithms, we constructed a 3-dimensional Swiss roll with holes, a standard test case for this task. Points were distributed in a plane with holes removed from the plane and the plane was wrapped in a spiral to create a relatively complex 3-dimensional structure. We obtained multiple views of the Swiss roll by projecting it onto planes at different angles, resulting in a number of affinity matrices forming a multi-view graph. The goal is to reconstruct the original Swiss roll geometry using the embeddings generated by each algorithm.
    
    \item \textbf{Stochastic Block Model}: 
    The stochastic block model (SBM) is used to simulate graphs with a latent cluster structure. To generate a graph under this model, the nodes are partitioned into $k$ equally sized clusters. Each possible edge is determined with a specific probability, with intra-cluster edges assigned with a probability of $\frac{c}{n}$, where $c$ represents the average degree of the graph; and inter-cluster edges assigned with probability $\frac{c(1-\lambda)}{n}$, where $\lambda$ is a parameter used to determine how distinct the clusters should be: $\lambda = 0.9$ implies that the edges are $\frac{9}{10}$ less likely to occur between clusters as within a cluster. To simulate multi-view graphs, we generate $m$ independent SBM graphs with the same set of parameters $n,k,c$, and $\lambda$, and the same partition of nodes for each view. Since this process does not necessarily produce connected graphs, we cull nodes that are not connected in each of the views. We measure clustering performance on this data set, varying the number of clusters, number of layers, and sparsity of the generated graphs.
\end{itemize}

\begin{figure*}[t]
    \centering
    \subfloat[CCR for varying number of views]{
        \includegraphics[width=.45\textwidth]{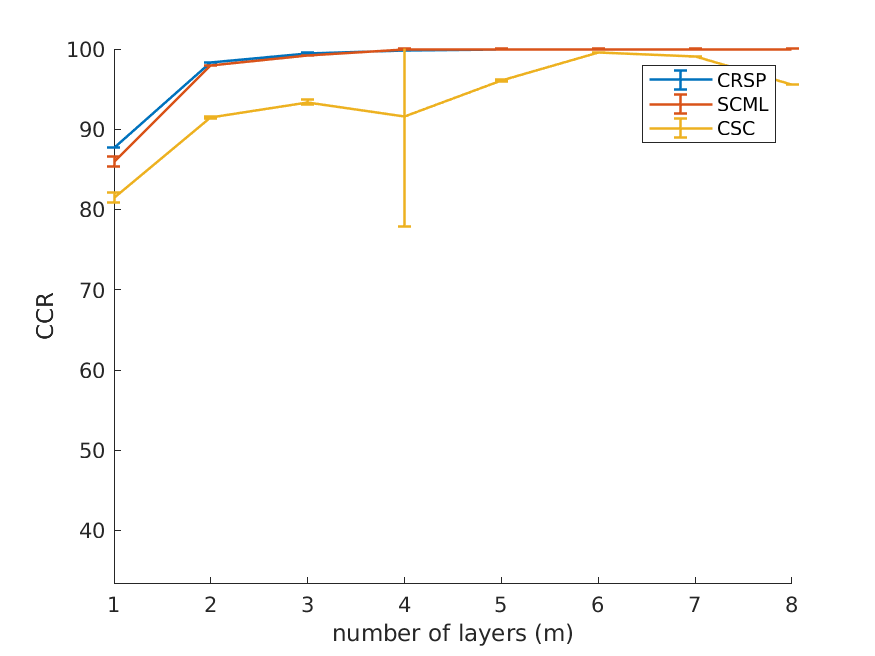}
        \label{fig:sbm_m_ccr}
    }
    \subfloat[NMI for varying number of views]{
        \includegraphics[width=.45\textwidth]{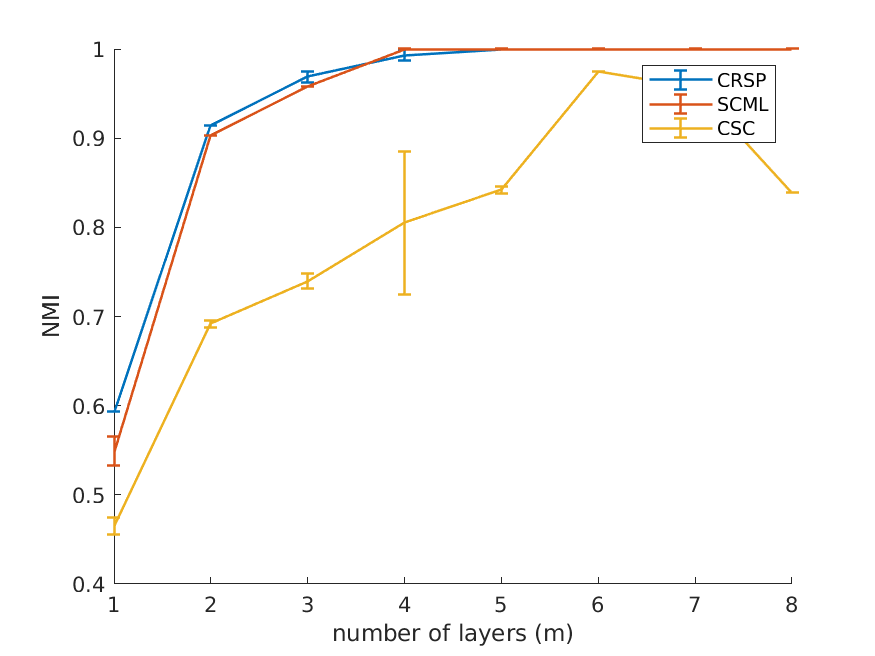}
        \label{fig:sbm_m_nmi}
    }
    \linebreak
    \subfloat[CCR for varying graph sparsity]{
        \includegraphics[width=.45\textwidth]{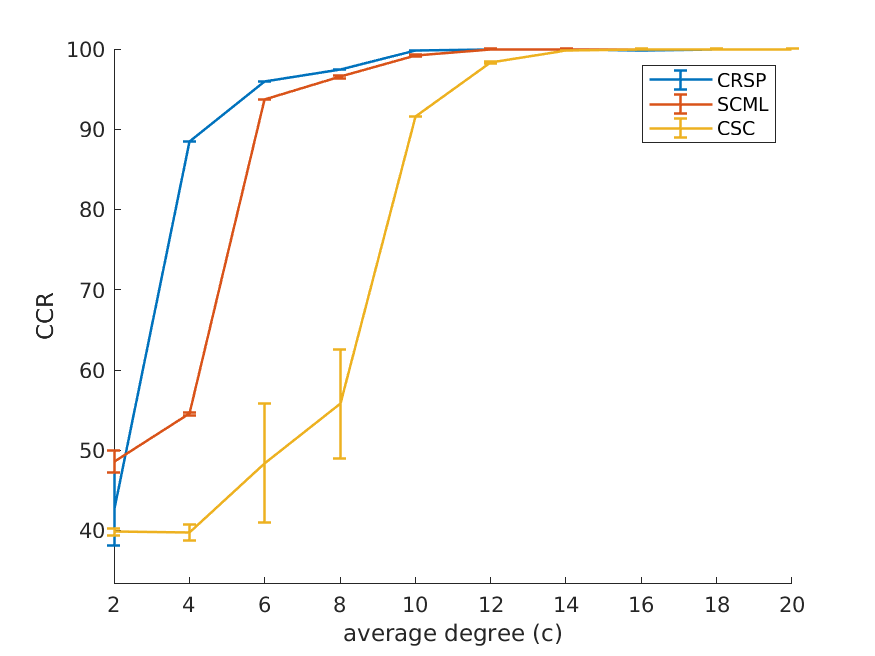}
        \label{fig:sbm_c_ccr}
    }
    \subfloat[NMI for varying graph sparsity]{
        \includegraphics[width=.45\textwidth]{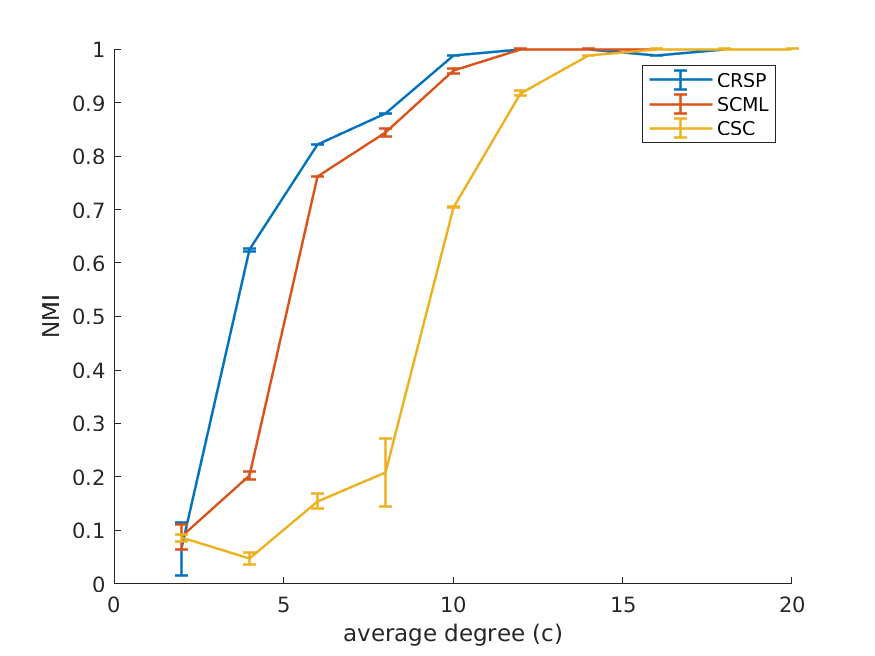}
        \label{fig:sbm_c_nmi}
    }
    \caption{Simulation results on Stochastic Block Model graphs with $n = 500, k = 3, \lambda = 0.9$ averaged over 10 runs. As the number of views increases, all algorithms perform better, with C-RSP outperforming the others. Similarly as the average degree increases, resulting in denser graphs, all algorithms perform more and more accurately, with C-RSP once again outperforming the two benchmark algorithms.}
    \label{fig:sbm}
\end{figure*}

\subsection{Real-world Data Sets}

\begin{itemize}
    \item \textbf{3Sources}\footnote{http://mlg.ucd.ie/datasets/3sources.htm}:
    This data set contains information about a set of news articles reported by three different news sources: the BBC, the Guardian, and Reuters. It covers 416 distinct news stories, of which 169 are reported on by all three agencies. These stories are classified under 6 disjoint clusters: business, entertainment, health, politics, sports, and technology. The three sources provide different views of the same news story, which are represented as different views of a multi-view graph. In our experiments, we extracted the 169 stories common to all three sources and constructed a $169\times 169$ affinity matrix for each source using a Gaussian kernel on the feature vectors provided. If $a_{ij}$ were the affinity between nodes $i$ and $j$ and their corresponding feature vectors were $x_i$ and $x_j$, then $a_{ij} = \exp\left(\frac{-\norm{x_i - x_j}^2}{2\sigma^2}\right)$, where $\sigma$ was taken to be the median of the pairwise Euclidean distances between the feature vectors. 

    \item \textbf{UCI Handwritten Digits}\footnote{https://archive.ics.uci.edu/ml/datasets/Multiple+Features}:
    In this data set, 2000 images of handwritten digits from 0 to 9 are analyzed, resulting in features matrix of Fourier coefficients, pixel averages, and several other feature matrices. In our experiments, we used 5 of the 6 feature types (excluding the Karhunen-Love coefficients) as our multiple views and constructed affinity matrices as before using a Gaussian kernel.
    
    \item \textbf{Multi-view Twitter}\footnote{http://mlg.ucd.ie/aggregation/index.html}:
    This data set consists of five Twitter user networks and various methods of interaction on Twitter. We chose the \emph{politics-uk} data set, consisting of 419 user accounts belonging to political figures and organizations from the UK and 3 views of their pairwise interactions: $x$ follows $y$,  $x$ mentions $y$, and $x$ retweets $y$. The user accounts form the nodes of the three $419\times 419$ graphs, and they are partitioned into 5 disjoint clusters based on political party affiliation: Labour, Conservative, Scottish National Party, Liberal Democrats, and other.
\end{itemize}

\begin{table*}[t]
    \centering
    \renewcommand{\arraystretch}{1.3}
    \caption{Clustering results on various real-world data sets, in the form ``Mean (Std)''}
    \begin{tabular}{|c|c|cccc|}
        \hline 
        \textbf{Metric} & \textbf{Data Set} & \textbf{C-RSP} & \textbf{SC-ML}  & \textbf{CSC}  & \textbf{MultiNMF} \\
        \hline 
        \hline
        \multirow{3}{4em}{CCR} 
        & \textbf{UCI}              & 86.96\% (4.10\%) & 80.09\% (6.76\%) & 82.11\% (7.99\%) & \cellcolor{gray!25} 92.33\% (0.03\%) \\
        & \textbf{3Sources}         & \cellcolor{gray!25} 58.22\% (4.14\%) & 51.48\% (3.55\%) & 43.55\% (2.59\%) & 34.50\% (0.02\%) \\
        & \textbf{MultiviewTwitter} & \cellcolor{gray!25} 82.84\% (3.96\%) & 69.86\% (1.50\%) & 49.54\% (3.19\%) & 56.67\% (0.01\%) \\
        \hline
        \multirow{3}{4em}{NMI}
        & \textbf{UCI}              & 0.80 (0.01) & 0.76 (0.03) & 0.78 (0.04) & \cellcolor{gray!25} 0.88 (0.02) \\
        & \textbf{3Sources}         & \cellcolor{gray!25} 0.56 (0.04) & 0.42 (0.02) & 0.31 (0.02) & 0.07 (0.01) \\
        & \textbf{MultiviewTwitter} & \cellcolor{gray!25} 0.60 (0.04) & 0.42 (0.01) & 0.28 (0.01) & 0.45 (0.01) \\
        \hline
    \end{tabular}
    \label{table:real_data}
\end{table*}

\subsection{Results on Synthetic Data}

We first tested the quality of the embeddings generated by C-RSP using the Swiss roll data set in figure 1. To obtain a multi-view graph from this data, the Swiss roll was rotated and projected into 2 dimensions, resulting in the four views pictured in figures \ref{fig:sr_v1} - \ref{fig:sr_v4}. An affinity matrix for each view was constructed using the pairwise Euclidean distances between points. In the case of C-RSP, the embeddings were generated from the output C-RSP distance matrix using Multidimensional Scaling. For the benchmarks, the embedding vectors generated via each algorithm prior to clustering the vectors were used to obtain labels for the nodes.

As seen in figure \ref{fig:sr_crsp}, the C-RSP embedding accurately captures the curvature of the Swiss roll and produces a slightly flattened version of the original spiral structure The embedding also retains the two holes present in the original Swiss roll data. The benchmark algorithms SC-ML, CSC, and MultiNMF all fail to recover the spiral structure and the holes, but retain the relative distances between nodes with some accuracy. This is shown by the grouping of similarly colored nodes in figures \ref{fig:sr_scml}-\ref{fig:sr_mnmf}. 

To evaluate C-RSP at clustering tasks, we tested it on synthetic multi-view graphs generated using the Stochastic Block Model. Unless otherwise noted, the graphs have $n = 500$ nodes, $k = 3$ clusters, an average node degree of $c=10$, and $\lambda = 0.9$. In figure \ref{fig:sbm}, we report the variation in CCR and NMI as the number of views of the multi-view graph increases, as well as the variation across multi-view graphs of different sparsity. In both cases, C-RSP shows significantly better clustering accuracy compared to the benchmark algorithms. 

\subsection{Results on Real-world Data}
Having observed that C-RSP shows high embedding and clustering accuracy on synthetic data sets, we now evaluate whether this performance holds on real-world data sets. For this, we run C-RSP and all the benchmark algorithms on three widely used multi-view data sets. In table \ref{table:real_data}, we report the CCR and NMI values obtained for each algorithm averaged over 10 runs with the standard deviation listed in parentheses.

C-RSP significantly outperforms the benchmark algorithms on the MultiviewTwitter data set with respect to both CCR and NMI. On 3Sources, the CCR of C-RSP is almost matched by SC-ML, but we observe a significant increase in the NMI of C-RSP compared to the other algorithms. MultiNMF shows much higher clustering performance on the UCI Handwritten Digits data set, while the other three methods have comparable values in both metrics. The dip in performance of C-RSP is likely due to the choice of parameter. For all experiments conducted above, we chose $\beta = 0.02$ for C-RSP since it is shown to be the optimal $\beta$ value for the RSP measure \cite{Kivimaki}. At this $\beta$, the C-RSP dissimilarities calculated tend more towards the commute time distances, which may not capture the graph structure effectively for graphs larger than 1000 nodes \cite{VonLuxburg:2014:HCT}, to which category the UCI data set falls. 

Overall, C-RSP provides superior clustering results, confirming that good multi-view graph embeddings results in higher clustering accuracy. Furthermore, the results show that C-RSP has robust performance across different types of data, providing reasonably high clustering across the board. A more extensive parameter study on C-RSP, which remains to be completed, would optimize the performance of C-RSP further. 

\subsection{Comparison of Computational Speed}

Since all algorithms were coded in MATLAB, we were able to obtain a fair comparison of their respective running times. On smaller graphs like 3Sources, C-RSP and SC-ML had comparable run times, both taking only 25\% of the time taken by CSC. All three algorithms were significantly faster than MultiNMF. On the larger UCI data set, SC-ML finished in roughly 60\% of the run time of C-RSP, but both were still faster than CSC and MultiNMF. Overall, C-RSP has an efficient run time, especially compared to CSC and MultiNMF, and we believe that it could be made more efficient by using faster method of computing the matrix inversion $(I - W)^{-1}$, which is the most computationally intensive step of the algorithm. A more rigorous experimental analysis of the comparative speeds of the multi-view algorithms remains to be completed. 

\section{Conclusion}

This paper introduced a novel distance measure for multi-view graphs named C-RSP (Common Randomized Shortest Paths), an extension of the RSP dissimilarity for single-view graphs. The C-RSP measure is a generalization of the commute time distance and the shortest path distance, which  allows it to encode both the local and global structure of a multi-view graph. This leads to more accurate graph embeddings, resulting in better visualization and high 
clustering accuracy. We tested C-RSP at both embedding and clustering tasks and showed that it produces superior results compared to other benchmark embedding algorithms while being computationally efficient.

\IEEEtriggeratref{10}

\bibliographystyle{IEEEtran}
\bibliography{biblio2}

\newpage

\appendix[Derivation of the C-RSP Distribution]

Suppose that $P_1^{\mathrm{ref}},\dots,P_m^{\mathrm{ref}}$ are the reference probability distributions of each view of a multi-view graph with cost matrices $C_1,\dots,C_m $. To obtain Common Randomized Shortest Paths (C-RSP), we solve the following optimization problem:

\begin{equation*}
\begin{split}
& P^{CRSP} = \argmin{P}{\sum_{i=1}^{m}\sum_{p\in\mathcal{P}_{s\to t}}P(p)C_i(p)} \\
& \text{subject to } \sum_{i=1}^m\sum_{p\in\mathcal{P}_{s\to t}}P(p)\ln\frac{P(p)}{P_i^{\mathrm{ref}}(p)} = J_0 \\
& \!\! \qquad \qquad \sum\limits_{p \in \mathcal{P}_{s\to t}}P(p) = 1
\end{split}
\end{equation*}

Consider a multi-view graph $\mathcal{G}$ with $m = 2$ layers, $G_1 = \{V, E_1\}$ and $G_2 = \{V,E_2\}$, with the reference transition probability distributions  $P_1^{\mathrm{ref}}$ and $P_2^{\mathrm{ref}}$ and cost matrices $C_1$ and $C_2$ respectively. To derive the common distribution (which we will call $Q$ for ease of notation) under the above stated constrained optimization, we use the following Lagrange function:

\begin{equation*}
\begin{split}
& \mathcal{L} = \sum_{i=1}^{m} \sum_{p\in\mathcal{P}_{s\to t}}Q(p)C_i(p) + \mu\left[\sum_{p\in\mathcal{P}_{s\to t}}Q(p) - 1 \right] \\
&+ \lambda\left[\sum_{p\in\mathcal{P}_{s\to t}}Q(p)\ln\frac{Q(p)}{P_i^{\mathrm{ref}}(p)} - J_0\right] 
\end{split}
\end{equation*}

Considering only one path, we obtain the following:

\begin{equation*}
\begin{split}
\frac{\partial\mathcal{L}}{\partial Q} &= \sum_{i=i}^m C_i(p) + \lambda\sum_{i=1}^m\left(\ln\frac{Q(p)}{P_i^{\mathrm{ref}}(p)} + 1\right) + \mu \\
&= \sum_{i=i}^m C_i(p) + \lambda\ln\left(\prod_{i=1}^m\frac{Q(p)}{P_i^{\mathrm{ref}}(p)}\right) + \lambda m + \mu \\
&= \sum_{i=i}^m C_i(p) + \lambda\ln\frac{Q^m(p)}{\prod_{i=1}^m P_i^{\mathrm{ref}}(p)} + \lambda m + \mu \\
&= 0
\end{split}
\end{equation*}

\newpage

or

\begin{equation*}
\ln\left[\frac{Q^m(p)}{\prod_{i=1}^m P_i^{\mathrm{ref}}(p)}\right] = -\frac{1}{\lambda}\sum_{i=1}^m C_i(p) - \frac{\mu}{\lambda} - m
\end{equation*}

which gives

\begin{equation*}
\begin{split}
Q^m(p) &= \prod_{i=1}^m P_i^{\mathrm{ref}}(p)\cdot e^{-\frac{1}{\lambda}\sum\limits_{i=1}^m C_i(p) - \frac{\mu}{\lambda} - m} \\
Q(p) &= \sqrt[m]{\prod_{i=1}^m P_i^{\mathrm{ref}}(p)}\cdot e^{-\frac{1}{m\lambda}\sum\limits_{i=1}^m C_i(p) - \frac{\mu}{m\lambda} - 1} \\
&= \sqrt[m]{\prod_{i=1}^m P_i^{\mathrm{ref}}(p)}\cdot \left[e^{-\frac{1}{m\lambda}\sum\limits_{i=1}^m C_i(p)}\right]\left[e^{- \frac{\mu}{m\lambda} - 1}\right] \\
&= c\sqrt[m]{\prod_{i=1}^m P_i^{\mathrm{ref}}(p)}\cdot e^{-\beta\sum_{i=1}^m C_i(p)} \\
&= c\,\overline{\mathbf{P}}\cdot e^{-\beta\overline{\mathbf{C}}}
\end{split}
\end{equation*}

Normalizing this to make it a probability distribution (which is the same as RSP, detailed in \cite{Kivimaki}), we obtain the following expression for the C-RSP probability distribution for a single path:
\begin{equation*}
P^{CRSP}(p_{s\to t}) = \frac{\sqrt[m]{\prod\limits_{i=1}^m P_i^{\mathrm{ref}}(p_{s\to t})}\cdot e^{-\beta\sum\limits_{i=1}^m C_i(p_{s\to t})}}{\sum\limits_{p\in\mathcal{P}_{s\to t}}\sqrt[m]{\prod\limits_{i=1}^m P_i^{\mathrm{ref}}(p)}\cdot e^{-\beta\sum\limits_{i=1}^m C_i(p)}}
\end{equation*}
When deriving the combined matrix $P^{CRSP}$, where all paths are considered, using the $P^{\mathrm{ref}}_i$ matrices, the multiplication must be done element-wise.  

Note that the constraint $J_0$ on the K\"ullback-Leibler Divergence disappears during the optimization and that it is not present in the expression derived for the C-RSP distribution above. Thus, $\beta := \frac{1}{m\lambda}$ is the only parameter that needs to be tuned for this distribution.

\end{document}